\documentclass[twoside,leqno,twocolumn]{article}

\usepackage[letterpaper]{geometry}
\usepackage{amsmath,amssymb,amsfonts}
\usepackage{algorithmic}
\usepackage{graphicx}
\usepackage{textcomp}
\usepackage{amssymb}
\usepackage{tabularx}
\usepackage{subcaption}
\usepackage{multirow}
\usepackage{ltexpprt}
\usepackage{hyperref}
\usepackage[table,xcdraw]{xcolor}
\begin{document}

\newcommand\relatedversion{}

\title{\Large VQA-GEN: A Visual Question Answering Benchmark for Domain Generalization\relatedversion}
\author{Suraj Jyothi Unni \thanks{School of Computing and Augmented Intelligence, Arizona State University. Email:\{sjyothiu, rmoraffa, huanliu\}@asu.edu}\and Raha Moraffah$^*$
\and Huan Liu$^*$}

\date{}

\maketitle


\fancyfoot[R]{\scriptsize{Copyright \textcopyright\ 2023 by SIAM\\
Unauthorized reproduction of this article is prohibited}}





\begin{abstract} \small\baselineskip=9pt Visual question answering (VQA) models are designed to demonstrate visual-textual reasoning capabilities. However, their real-world applicability is hindered by a lack of comprehensive benchmark datasets. Existing domain generalization datasets for VQA exhibit a unilateral focus on textual shifts while VQA being a multi-modal task contains shifts across both visual and textual domains. We propose VQA-GEN, the first ever multi-modal benchmark dataset for distribution shift generated through a shift induced pipeline. Experiments demonstrate VQA-GEN dataset exposes the vulnerability of existing methods to joint multi-modal distribution shifts. validating that comprehensive multi-modal shifts are critical for robust VQA generalization. Models trained on VQA-GEN exhibit improved cross-domain and in-domain performance, confirming the value of VQA-GEN. Further, we analyze the importance of each shift technique of our pipeline contributing to the generalization of the model.
\end{abstract}
\section{Introduction}
Visual question answering (VQA) has emerged as a canonical task for evaluating multi-modal intelligence, requiring joint understanding of visual and textual inputs. Significant progress has been made in VQA, with models achieving human-level performance on many standard benchmarks. However, robustness remains a central concern for deploying VQA systems to real-world applications.\\
A core assumption in machine learning is that training and test data are drawn from the same underlying distribution. However, with distributions shift across domains model performance often degrades sharply. While domain generalization datasets expose models to varied training distributions, constructing comprehensive benchmarks spanning the full multi-modal distribution shifts remains an open challenge for VQA.\\
Existing benchmark Domain Generalization datasets such as VQA-CP \cite{DBLP:journals/corr/abs-1712-00377}, and VQA-Compose \cite{DBLP:journals/corr/abs-2002-08325} focus on textual shift but multimodal datasets have shifts in both text and image modalities. As a result, models trained on existing domain generalization data exhibit limited real-world effectiveness due to single modality focuses.\\
Currently, no VQA datasets incorporate coordinated multimodal shifts across images and text. Techniques that can systematically produce coherent variations in both modalities are lacking.In this paper, we present VQA-GEN - a novel large-scale dataset for multi-modal distribution shifts.\\
\begin{table}[t]
\centering
\begin{tabular}{lcc}
\hline
\hline
\textbf{Dataset} & \textbf{Image Shift} & \textbf{Question Shift} \\
\hline
VQA-GEN & \checkmark & \checkmark \\
VQA-Compose & $\times$ & \checkmark \\
VQA-CP & $\times$ & $\times$ \\
\hline
\end{tabular}
\caption{Domain Generalization Datasets Comparison}
\label{tab:datasets}
\end{table}
VQA-GEN is constructed through an intricately designed three-stage data generation pipeline for inducing controlled multi-modal shifts. The first stage introduces comprehensive visual shifts in the input images through techniques like style transfer and image corruptions, generating stylistic and perturbed variants. The second stage elicits linguistic variations in the source questions using methods such as backward translation and persona-based modeling to induce textual shifts synthetically. The final stage recombines the shifted images and questions through a mix-and-match process to create novel paired cross-modal distributions- the essence of VQA-GEN. This automated pipeline ensures that the contextual alignment of images, questions, and answers remains consistent with the original VQA data. A comparative analysis between existing domain generalization datasets and the proposed VQA-GEN dataset is presented in Table 1.\\
\begin{figure*}
  \centering
  \includegraphics[height=7cm,width=14cm]{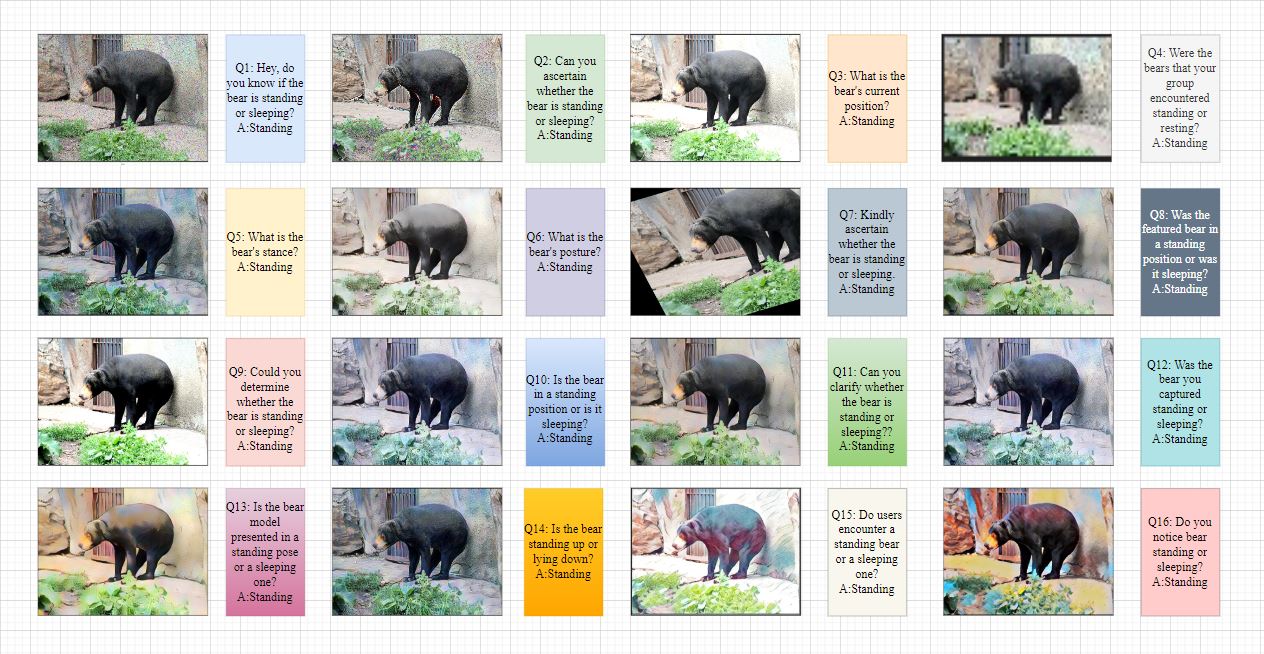}
 \caption{Example of VQA-GEN datasets showcasing image shifts of standing bear, accompanied by corresponding question shifts inquiring about its posture (standing or sleeping).}
\end{figure*}
Experiments demonstrate that the VQA-GEN dataset exposes the limitations of existing methods when faced with joint multimodal distribution shifts. Furthermore, models trained on VQA-GEN substantially outperform those trained on existing domain generalization datasets on out-of-domain target tasks. This highlights the usefulness of VQA-GEN for training robust models under multimodal shifts. Additional in-dataset evaluations validate the efficacy of each shift technique of our pipeline for evaluating generalized VQA models. We also analytically show the quality of the generated images and questions through domain shift analysis and similarity metrics.Figure 1 shows example of VQA-GEN dataset.\\
\begin{table*}[]
\centering
\begin{tabular}{lllp{3cm}p{3cm}p{3cm}}
\hline
\hline

Models & VQA-Compose & VQA-CP & VQA-GEN(Joint Shifts) & VQA-GEN(Image Shifts) & VQA-GEN(Question Shifts) \\
\hline
ViLT   & 55.82 $\pm$ 1.21 & 67.91 $\pm$ 1.37 & 42.26 $\pm$ 0.08 & 50.24 $\pm$ 2.25 & 47.86 $\pm$ 0.85 \\
MAC    & 38.65 $\pm$ 0.6 & 50.35 $\pm$ 0.3 & 31.95 $\pm$ 0.7 & 36.82 $\pm$ 0.8 & 37.63 $\pm$ 0.6 \\
RelNet & 58.82 $\pm$ 0.8 & 64.27 $\pm$ 0.7 & 37.91 $\pm$ 0.9 & 45.62 $\pm$ 0.6 & 43.82 $\pm$ 0.7 \\
\hline
\end{tabular}
\caption{Comparison of Validation Accuracy (\%) among ViLT, MAC, and RelNet models on different VQA datasets (VQA-Compose, VQA-CP, VQA-GEN(Joint Shift), VQA-GEN(Image Shift), VQA-GEN(Question Shift)). The values represent the model's accuracy.}
\label{tab:generalization_comparison}
\end{table*}
\begin{table}[ht]
    \centering
    \begin{tabular}{p{3cm}p{3cm}}
    \hline
    \hline
    \textbf{Dataset Splits} & Train/Valid/Test (Same as VQA v2) \\ \hline
    \textbf{Question Length} & 5 to 15 words \\ \hline
    \textbf{Vocabulary Size} & 82,000 unique words \\ \hline
    \textbf{QA Pairs} & Over twenty-three million \\ \hline
    \end{tabular}
    \caption{Statistics of the VQA-GEN Dataset}
    \label{tab:vqa-gen-stats}
\end{table}
Together, these results validate that aligned multimodal shift is imperative for VQA robustness. VQA-GEN fills this need through controlled visual and textual shifts beyond current domain generalization VQA datasets.\\
In summary, this paper contributes in the following ways:
\begin{enumerate}
 \item We propose VQA-GEN, the \textit{first} comprehensive benchmark dataset for domain generalization in VQA encompassing multi-modal shifts through controlled visual and textual variations. 
  \item We present a multi-modal pipeline for generating coherent shifts that are logically aligned across image and text domains.
 \item We demonstrate that comprehensive multi-modal shift is pivotal for achieving robust VQA generalization.

\end{enumerate}

\section{Related Work}
\textbf{Domain generalization} datasets in the context of Visual Question Answering (VQA) have garnered significant attention in research. While the VQA-CP approach divides data based on the answer distribution of the VQA v2 dataset, it fails to adequately consider crucial factors such as image and question variations. Conversely, the VQA Compose method primarily focuses on logically composed binary questions, overlooking the broader spectrum of question types. Here, we propose VQA-GEN dataset, which aims to address both image and question bias in the domain generalization of VQA.\\\\
\textbf{Text-based shift techniques} have been extensively investigated to generate meaningful variations in text.SimpleAug \cite{DBLP:journals/corr/abs-2109-06122}, is one such technique that relies on paraphrasing but lacks diversity within the dataset. Another approach, DoCoGen \cite{calderon2022docogen}, utilizes the T5 model \cite{raffel2020exploring} to generate text variations. However, studies have shown that the T5 model often deviates from the original syntax, resulting in a small proportion of generated sentences that closely align with the source text. Other approaches such as multitask learning \cite{niu-etal-2018-multi}, reinforced learning \cite{sancheti2020reinforced}, and augmenting pre-trained models with rewards \cite{lai-etal-2021-thank} , require targeted text.\\ 
In the realm of unsupervised approaches, techniques like  disentangling style and content \cite{shen2017style}, style-word editing \cite{wu-etal-2019-hierarchical-reinforced}, utilizing generic resources \cite{chawla-yang-2020-semi}, ChatGPT-based style transfer \cite{lai2023multidimensional}, back-translation \cite{junczys-dowmunt-etal-2018-marian}
have been investigated. For our pipeline, we employed back-translation and style transfer techniques using ChatGPT with different personas to generate controlled question shifts.\\\\
\textbf{Image-based shift techniques} encompass a range of methods used to modify the position and appearance of an image within its background. One commonly used technique is masking, where specific objects or regions within an image are hidden to introduce variations. However, masking can introduce bias and result in information loss, as it overly emphasizes question-based shifts while neglecting other important visual features and variations.\\
Another widely studied technique is image style transfer, which allows the generation of artistic paintings without requiring the expertise of a professional painter. A study by \cite{7780634} found that the inner products of feature maps in Convolutional Neural Networks (CNNs) can represent styles, leading to the proposal of a neural style transfer (NST) method through successive optimization iterations. However, this optimization process is time-consuming and challenging to widely apply.\\
Furthermore, techniques such as Imagenet-C \cite{hendrycks2019benchmarking} and Imagenet-P perturbations are utilized to generate image shifts. Approaches [ \cite{huang2017arbitrary}, \cite{li2017universal}] focus on aligning the second-order statistics of style and content images to enable arbitrary style transfer. One approach presented by \cite{huang2017arbitrary} employs adaptive instance normalization (AdaIN) to normalize content features using the mean and variance of style features, facilitating arbitrary style transfer. Another approach, StyTr2  \cite{deng2022stytr2}, incorporates separate transformer encoders to produce domain-specific sequences for content and style. Additionally, Contrastive Arbitrary Style Transfer (CAST) \cite{Zhang_2022} considers style distribution and enables users to learn style representations directly from image attributes.\\
In our approach, we manually selected style images that preserve the semantics of the images and utilized the technique proposed \cite{huang2017arbitrary}, along with Imagenet-C transformations, to generate shifts in images.
\section{Preliminary Assessment}
In this section, we conduct a preliminary assessment to validate the need for multi-modal shifts in exposing model vulnerabilities toward robustness. We evaluated three state-of-the-art VQA models from different categories: classic two-stream (RelNet), transformer (ViLT), and hybrid neuro-symbolic (MAC). The models are trained on VQA v2 and tested on VQA-CP, VQA-GEN (Image shifts only), VQA-GEN (Question shifts only), and VQA-GEN (Joint Shifts).\\
The results from Table 2 yield critical insights. Firstly, while performance on VQA-CP (50.35-67.91\%) and VQA-Compose (55.82-58.82\%) suggests current VQA models have good generalizability for textual shifts, their accuracy still drops substantially to 43.82-47.86\% on VQA-GEN (Question Shifts). This reveals existing models are not fully robust to textual shifts.\\
Secondly, the precipitous decline in accuracy from 50.35-67.91\% on VQA-CP to 36.82-50.24\% on VQA-GEN (Image Shifts) highlights models' high vulnerability to image shifts. As existing datasets focus solely on textual shifts, models are never exposed to, and thus unequipped to handle, real-world image shifts.\\
Most critically, joint shifts in VQA-GEN cause a huge performance dip, with accuracies plummeting from 50.35-67.91\% on VQA-CP to 31.95-42.26\% on VQA-GEN (Joint Shifts). This exposes critical gaps missed by single shifts. Using this joint shift approach allows VQA-GEN to comprehensively evaluate model robustness by introducing multi-modal
variations that reveal vulnerabilities.
\begin{figure*}[!ht]
  \centering
  \includegraphics[height=5cm,width=13cm]{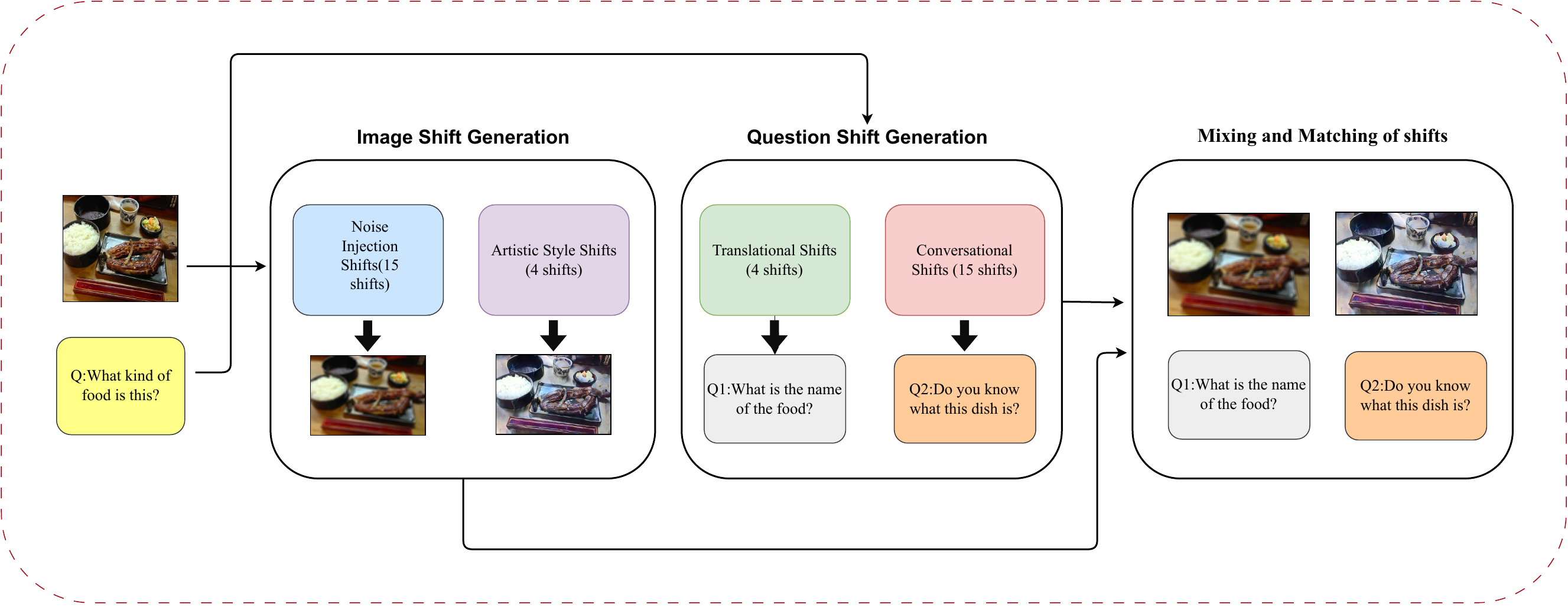}
  \caption{An overview of our pipeline. We first generate image shifts to the image input through Noise and Style shifts and the input question is passed to the question shift module which generates diverse questions using translational and conversational shift techniques. Finally, the generated shifts from each modal are combined using mix and match, resulting in new image question pairs which are shown on the right.}
\end{figure*}
variations that reveal vulnerabilities.
\section{VQA-GEN Dataset}
Table 3 provides an overview of VQA-GEN dataset statistics. In this section, we discuss about VQA-GEN dataset generation process. Figure 2 shows the data generation pipeline to construct VQA-GEN. The pipeline consists of three main steps: (1) generating image shifts, (2) generating question shifts, and (3) mixing and matching the shifts into new image-question pairs. Although we could have used any VQA dataset, we leverage VQA v2 as the source dataset for its wide acceptance and real-world-based images.
\subsection{Generating Image Shift}
Our aim was to generate image shifts that preserve the essential features like shape, color, etc. required for addressing the given question. Through our exploration, we found three techniques for creating shifts in images: masking, style transfer, and noise perturbations. Out of these three, we found masking to be less effective as there is a higher chance of losing major features associated with the questions using existing masking techniques.
Consequently, we focused on two approaches: (1) Noise Injection Shifts and (2) Artistic Style Shifts.
\begin{figure}[t]
  \centering
  \includegraphics[height=5cm,width=7cm]{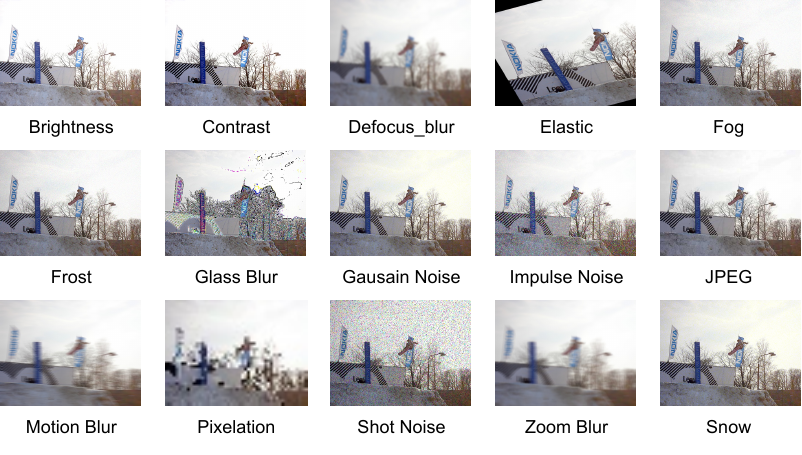}
 \caption{Perturbations methods in Noise Injection Shifts}
\end{figure}
\subsubsection{Noise Injection Shifts:}
In our first technique, we draw inspiration from the ImageNet-C dataset \cite{huang2017arbitrary}. This image generation process ensures controlled perturbations that retain critical objects and features. The specific corruptions used for our pipeline, as shown in Figure 3, include:
\textbf{Gaussian noise} adds subtle pixel-level fluctuations generated by sampling from a normal distribution.
\textbf{Shot noise} results in a grainy appearance akin to low-light photography interference via a Poisson distribution.
\textbf{Impulse noise} introducing pixel anomalies through a Bernoulli distribution.
Blur effects of image like \textbf{defocus and motion blur} using convolutional filters.Visual effects simulating atmospheric conditions like \textbf{snow, frost, and fog} based on procedural noise generation.
\textbf{brightness and contrast shifts}  adjust image brightness and differences between light and dark areas using uniform distributions. Stylistic blur techniques like \textbf{glass and zoom blur} using FFTs.
\textbf{Pixelation}  transform images into blocky forms through spatial binning.
\textbf{JPEG compression} artifacts through encoding.\textbf{Elastic warping distortions} for elastic and non-linear image deformations via spline grids.
Using this technique, we create 15 perturbations for a single input image.

\subsubsection{Artistic Style Shifts:} In the second technique, we leveraged the technique of Adaptive Instance Normalization (AdaIN) \cite{huang2017arbitrary}. AdaIN transfers styles by normalizing feature maps from the content image to match the mean and variance statistics of feature maps from the style image. This alignment of feature statistics allows the network to stylize the content image with the artistic style of the style image. Based on our experiments, we found 4 style shifts - dramatic, art, watercolor, and cartoon - that provide significant artistic diversity while avoiding surreal/abstract styles that could distort semantic content needed for VQA.\\
The culmination of these noise and artistic style shifts yields a comprehensive set of 19 image cues per original input image.
\begin{figure}
  \centering
  \includegraphics[height=4cm,width=7cm]{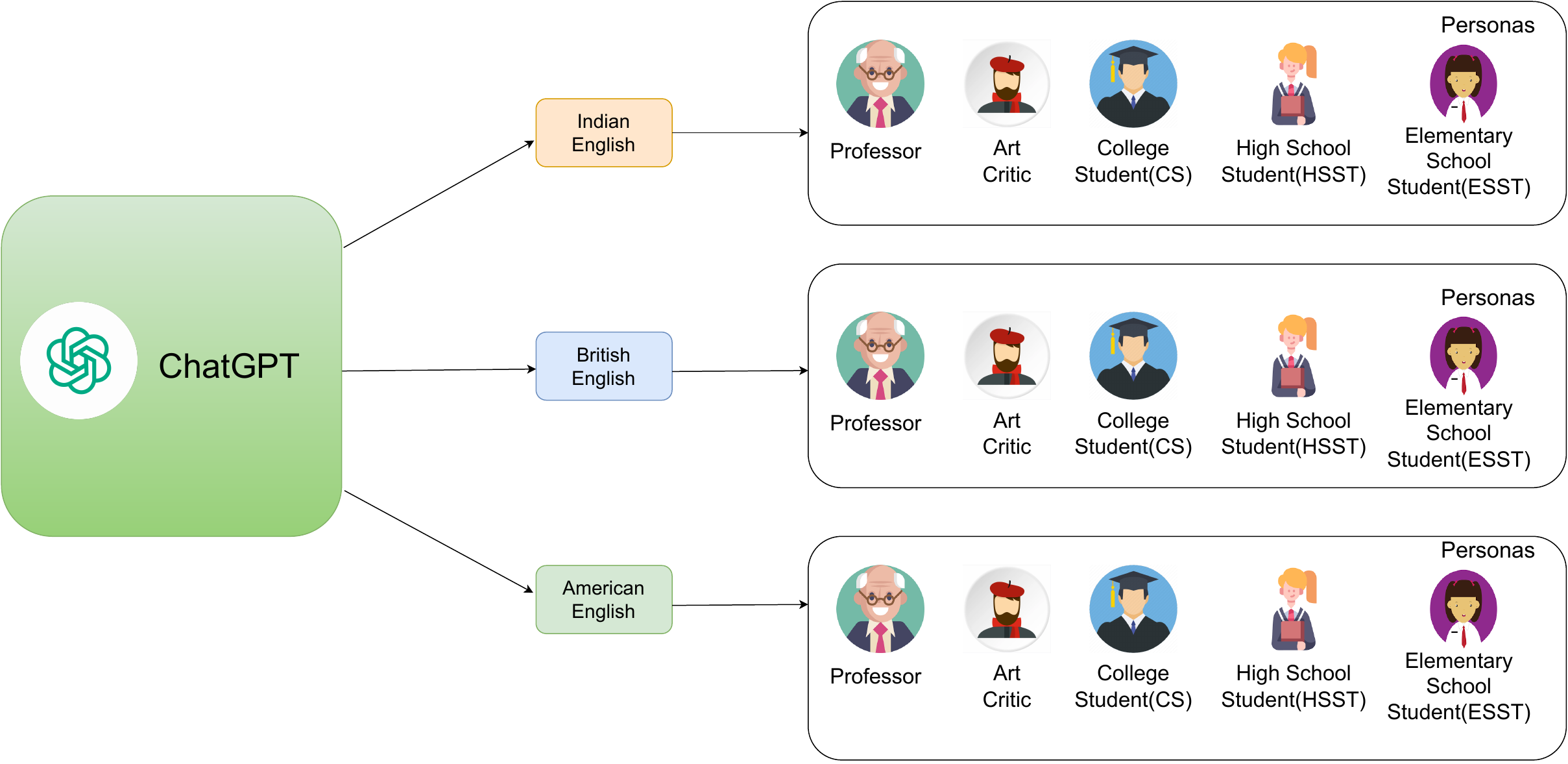}
 \caption{Persona-Based Question Shift}
\end{figure}
\subsection{Generating Question Shift}
The majority of text shifts explored in our research were focused on employing paraphrasing techniques. Although reasoning-based questions and logical compositions were found to be effective, they often generated different answers, which did not align with the requirements of our pipeline. Our pipeline, on the other hand, aimed to maintain a fixed answer while introducing variations in the image and question shifts. Through our investigation, we came up with two techniques aligning with controlled shift generation: (1) Translational Shifts (TS) and (2) Conversational Shifts (CS).

\subsubsection{Translational Shifts (TS):} In the first technique, we translate questions to French and back to English using MarianNMT, a machine translation framework based on the Transformer architecture for generating novel text cues. We chose French as an intermediary language because its grammatical and structural differences from English can elicit greater linguistic diversity. Through human evaluation of translations on 10 random questions, we found MarianNMT generated 8 semantically similar variations per question on average. This significantly outperformed paraphrasing models like T5, which struggled to preserve semantics. By leveraging the capabilities of MarianNMT translation paired with French's linguistic divergence, we generate 4 variations per original question.
\subsubsection{Conversational Shifts (CS):} The second technique generate shifts by emulating distinct English writing styles prevalent in India, Britain, and the United States, we further categorized these styles based on personas, such as high school student (HSST), critic, professor, elementary school student (EST), and college student (CS) as shown in Figure 4. To create these diverse styles, we harnessed ChatGPT \cite{bishop2023computer}, a large language model (LLM) adept at simulating various writing styles. By providing ChatGPT with hard prompts specifying a particular persona, we are able to elicit 15 distinct stylistic variations for each original question. In total, 19 textual cues are generated per input question through the combination of translational and conversational shifts.
\subsection{Mixing and Matching}
The image shifts and question shifts generated from the previous steps are combined in the final step. Since the generated images and questions share the same context, we randomly associate questions with images, resulting in 19 different distributions for a single input image and question combination. This approach is consistently applied to all image-question pairs of VQA v2 to generate VQA-GEN dataset. The modular architecture of the pipeline enables its independent application to various VQA datasets, facilitating the conversion into a Domain-Generalized VQA (DG VQA) format.
\begin{table*}[!ht]
\centering
\begin{tabular}{lclcllcllcllcllcllclc}
\hline
\hline
Datasets & \multicolumn{2}{l}{VQA-GEN} & \multicolumn{3}{l}{VQA-CP} & \multicolumn{3}{l}{VQA-Compose} & \multicolumn{3}{l}{CLEVR} & \multicolumn{3}{l}{GQA Bal} & \multicolumn{3}{l}{VQA Abs} & \multicolumn{3}{l}{Visual Genome} \\
\hline
VQA-GEN & \multicolumn{2}{c}{\cellcolor{blue!25}\textbf{64.2}} & \multicolumn{3}{c}{\textbf{75.6}} & \multicolumn{3}{c}{\textbf{68.4}} & \multicolumn{3}{c}{\textbf{58.9}} & \multicolumn{3}{c}{\textbf{53.9}} & \multicolumn{3}{c}{\textbf{52.2}} & \multicolumn{3}{c}{\textbf{54.8} }\\
VQA-CP & \multicolumn{2}{c}{30.3} & \multicolumn{3}{c}{\cellcolor{blue!25}66.4} & \multicolumn{3}{c}{39.4} & \multicolumn{3}{c}{31.0} & \multicolumn{3}{c}{33.6} & \multicolumn{3}{c}{33.1} & \multicolumn{3}{c}{29.7} \\
VQA-Compose & \multicolumn{2}{c}{32.1} & \multicolumn{3}{c}{63.1} & \multicolumn{3}{c}{\cellcolor{blue!25}84.4} & \multicolumn{3}{c}{42.1} & \multicolumn{3}{c}{46.4} & \multicolumn{3}{c}{37.2} & \multicolumn{3}{c}{39.5} \\
CLEVR & \multicolumn{2}{c}{19.6} & \multicolumn{3}{c}{25.7} & \multicolumn{3}{c}{21.3} & \multicolumn{3}{c}{\cellcolor{blue!25}84.9} & \multicolumn{3}{c}{21.4} & \multicolumn{3}{c}{21.7} & \multicolumn{3}{c}{23.1} \\
GQA Bal & \multicolumn{2}{c}{30.2} & \multicolumn{3}{c}{38.3} & \multicolumn{3}{c}{32.6} & \multicolumn{3}{c}{36.8} & \multicolumn{3}{c}{\cellcolor{blue!25}58.2} & \multicolumn{3}{c}{39.8} & \multicolumn{3}{c}{30.6} \\
VQA Abs & \multicolumn{2}{c}{35.7} & \multicolumn{3}{c}{47.2} & \multicolumn{3}{c}{40.5} & \multicolumn{3}{c}{41.2} & \multicolumn{3}{c}{38.2} & \multicolumn{3}{c}{\cellcolor{blue!25}56.3} & \multicolumn{3}{c}{41.3} \\
Visual Genome & \multicolumn{2}{c}{31.5} & \multicolumn{3}{c}{46.5} & \multicolumn{3}{c}{39.2} & \multicolumn{3}{c}{32.1} & \multicolumn{3}{c}{34.4} & \multicolumn{3}{c}{35.2} & \multicolumn{3}{c}{\cellcolor{blue!25}41.0} \\
\hline
\end{tabular}
\caption{Validation accuracy (\%) of VQA datasets on ViLT model. Blue shading represents in-domain accuracy, highlighting the diagonal cells where each dataset's performance on itself is shown for reference.The remaining cells indicate cross-domain accuracy.}
\end{table*}
\section{Experiments}
This section discusses the experimental settings used to validate the need for the VQA-GEN dataset, including the datasets and evaluation metrics used, and the baselines, followed by a detailed analysis of the experiments. We conduct a series of experiments to answer the following research questions.
\begin{enumerate}
    \item[--] \textbf{RQ.1:} Does the identified multi-modal dataset improve the in-domain and cross-domain performance of the models?
    \item[--] \textbf{RQ.2:} What is the importance of each VQA-GEN generation pipeline shift technique in improving the generalization performance?
    \item[--] \textbf{RQ.3:} Does VQA-GEN provides diverse variations while maintaining relevance for improving VQA models?
\end{enumerate}

\subsection{Baseline Datasets}
COCO-QA (120k images, 330k QA pairs) focuses on VQA with diverse images and question-answer sets. Visual Genome (108k images, 1.7M QA pairs) provides extensive image annotations for scene understanding. CLEVR (100k synthesized images, 1M QA pairs) evaluates reasoning with complex questions about 3D scenes. Visual7W (47k images, 327k QA pairs) tests natural language understanding and reasoning. VQA Abstract (50k abstract scenes, 150k QA pairs) challenges models with abstract images. VizWiz (31k real images, 160k QA pairs) addresses real-world VQA for blind users. VQA v2 (204k COCO images, 1.1M QA pairs) provides a VQA benchmark with additional data. VQA-CP (204k COCO images, 2.8M QA pairs) tests generalization with counterfactual questions.
\subsection{Evaluation Metrics}
We use these metrics for the evaluation of our experiment:
\textbf{Accuracy :}
We calculate top-1 accuracy on the validation split for each dataset by matching predictions to ground-truth answers. This evaluates model performance on the variations.
\textbf{MMD Loss :}
We utilize the Maximum Mean Discrepancy (MMD) metric for domain shift analysis, a statistical test measuring the difference between probability distributions. In this, we use a pre-trained ResNet-101 model to extract visual features, encompassing high and low-level information about the image. In the question space, BERT encodes 10,000 questions to capture semantic content, and we extract 20 syntactical features like question length. By calculating the MMD based on these features, we quantify distribution shift, with lower MMD values indicating greater similarity between distributions.
\textbf{Similarity Analysis :}
To compare the original dataset with VQA-GEN, we employed similarity metrics like BLEU-1, BLEU-2, BLEU-3, BLEU-4, and METEOR to assess the quality and diversity of the generated questions.
\textbf{BLEU} (Bilingual Evaluation Understudy) measures similarity based on overlapping n-grams between a candidate and reference text. Higher BLEU scores indicate greater similarity.
\textbf{METEOR} (Metric for Evaluation of Translation with Explicit Ordering) calculates similarity using explicit word-to-word matches, with recall weighted higher than precision.
\subsection{Implementation Details}
In our experiments, we utilized the ViLT architecture for our training process. Our models underwent 100 epochs of training on 2 Nvidia Titan X GPUs, employing the Adam optimizer and a batch size of 64.\\
We created custom splits for VQA-GEN to evaluate image and question shifts. We had a training split for image shifts with 10 types of noise shifts (Gaussian Noise, Shot Noise, etc.) applied to 400,000 QA pairs. The test split had 5 additional noise shifts (Glass Blur, JPEG artifacts, etc.) applied to 200,000 QA pairs. We also had a training split with 2 style shifts (Dramatic, Art) applied to 400,000 QA pairs, and a test split with 2 additional style shifts (Watercolor, Cartoon) applied to 200,000 QA pairs. For question shifts, the training split contained 400,000 QA pairs with translational (trans.) shifts while the test split had 200,000 QA pairs with translational shifts. Additionally, the conversational(conv.) shift training split had 3 personas (High School Student, Critic, Professor) applied to 400,000 QA pairs, and the test split had 2 more personas (Elementary Student, College Student) applied to 200,000 QA pairs.
\begin{table*}[ht]
\centering
\renewcommand{\arraystretch}{1.5} 
\setlength{\tabcolsep}{6pt} 
\begin{tabular}{>{\raggedright}p{6cm} >{\raggedright}p{6cm} l}
\hline
\hline
\textbf{Training Splits} & \textbf{Test Splits} & \textbf{Validation Accuracy} \\
\hline
Noise Shifts & Noise Shifts Test & $69.3 \pm 1.2$ \\
Style Shifts & Style Shifts Test & $67.1 \pm 0.8$ \\
Trans. Shifts & Trans. Shifts Test & $66.8 \pm 1.4$ \\
Conv. Shifts & Conv. Shifts Test & $65.4 \pm 2.3$ \\
Noise Shifts + Trans. Shifts  & Noise Shifts Test + Trans. Shifts  Test & $62.5 \pm 0.6$ \\
Noise Shifts + Conv. Shifts & Noise Shifts Test + Conv. Shifts & $63.7 \pm 1.1 $ \\
Style Shifts + Trans. Shifts & Style Shifts Test + Trans. Shifts Test & $63.2 \pm 0.3$ \\
Style Shifts + Conv. Shifts & Style Shifts Test + Conv. Shifts Test & $62.9 \pm 0.2$ \\
\hline
\end{tabular}
\caption{Validation Accuracy (\%)  of ViLT Model on VQA-GEN Custom Splits(refer implementation details section).}
\label{table:in-dataset}
\end{table*}
\subsection{RQ.1 Performance Comparison}
In this section, we evaluate the cross-domain and in-domain generalization capabilities of the VQA-GEN dataset. In-domain refers to training and evaluating on the same dataset, while cross-domain involves training on one dataset and evaluating on another. Table 4 shows the comparison results of VQA-GEN and the existing VQA datasets. Here, the ViLT model was trained and evaluated on: VQA-GEN, VQA-Compose, VQA-CP, CLEVR, VizWiz, VQA Abstract, and Visual Genome. We observe that VQA-GEN dataset proves to be useful source data when transferring knowledge to other VQA datasets, achieving the best results on VQA-CP with a validation accuracy of 75.6\% and least for VQA Abs. which is around 52\%. We also noted that existing domain generalization datasets like VQA-CP and VQA-Compose do improve the model's performance on some cross-domain datasets. For VQA-CP, it scores lower than VQA-GEN on 6 out of 7 cross-domain datasets, with accuracies in the 30-47\% range. Meanwhile, VQA-Compose achieves the top accuracy on CLEVR (42.1\%) but poorer performance on VQA Abstract (37.2\%) and Visual Genome (39.5\%). This suggests while some domains transfer well, others exhibit a larger gap. Apart from that, the CLEVR dataset appears to have the least impact on cross-domain tasks, yielding accuracy rates of approximately 20-25\% across all datasets.\\
Furthermore, we investigated in-domain generalization, represented along the left diagonal in Table 3. VQA-GEN attains 64.2\% validation accuracy, indicating that ViLT successfully learns from its training set. However, this is lower than existing domain generalization datasets like VQA-CP (66.4\%) and VQA-Compose (84.4\%). In-domain validation accuracy is lowest for Visual Genome, at 41.0\%, suggesting that the model doesn't learn as much from its training split. We can see that domain generalization datasets proved to be effective for the model by providing a wide range of shifts that enable the model to adapt effectively. Notably, VQA-GEN distinguishes itself due to its unique composition, encompassing joint shifts in both image and question domains and also validating the RQ.1.

\subsection{RQ.2 In-dataset Generalization}
Table 5 shows the effectiveness of each shift technique in the VQA-GEN data generation pipeline which we call In-dataset generalization. Here we determine whether the performance gains of VQA-GEN result from the applied shifts or the expanded scale of the dataset. ViLT model was trained on each training split and evaluated on testing splits of VQA-GEN mentioned in the implementation detail section. We also combine training splits and evaluate combined test splits to analyze the effect of modeling diverse shifts. We find that noise injection shifts achieve the highest accuracy of 69.3\%, indicating the model is most robust to visual corruptions. Artistic style transfers (67.1\%) also provide relatively effective generalization, likely due to the model's ability to adapt to stylistic image changes. However, textual shifts like translation (66.8\%) and conversational shifts (65.4\%) prove more challenging. The model may struggle to handle semantic variations in questions generated through these approaches. Combining training splits leads to lower accuracy than individual shifts. For instance, accuracy drops to 62.5\% when jointly modeling noise injection and translational shifts. This highlights the difficulty in adapting to the full diversity of shifts in VQA-GEN. We found that all these shift techniques provide a challenging evaluation dataset to assess the model. Also on the joint shift data, we observe the model's performance drops significantly to the unimodal shifts proving the usefulness of the dataset for training robust model(RQ2).
\begin{figure*}
  \centering
  \begin{subfigure}{0.5\textwidth}
    \centering
    \includegraphics[width=\linewidth]{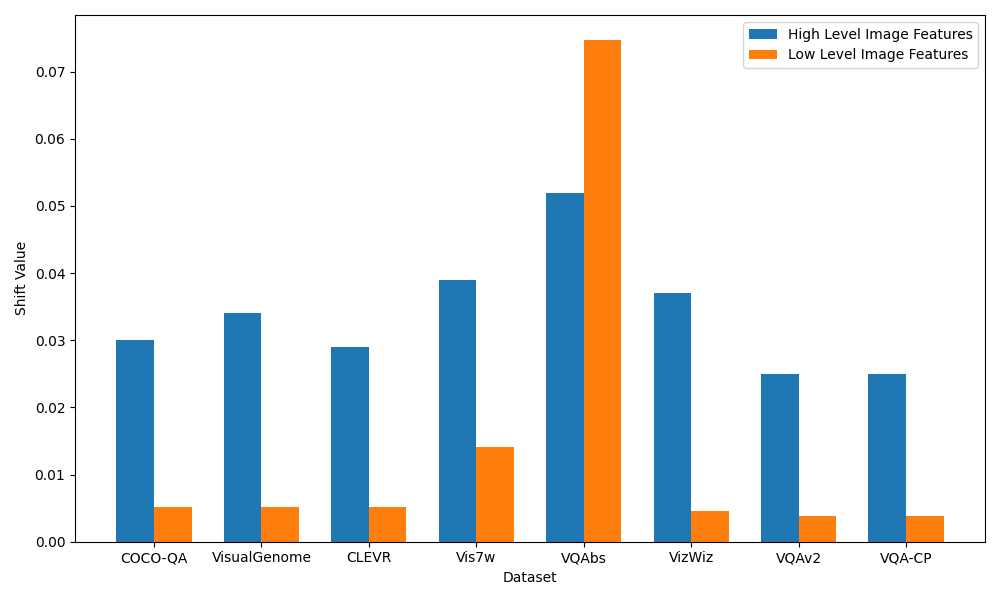}
    \caption{High and low level Image feature shifts.}
    \label{fig:image-shift}
  \end{subfigure}%
  \begin{subfigure}{0.5\textwidth}
    \centering
    \includegraphics[width=\linewidth]{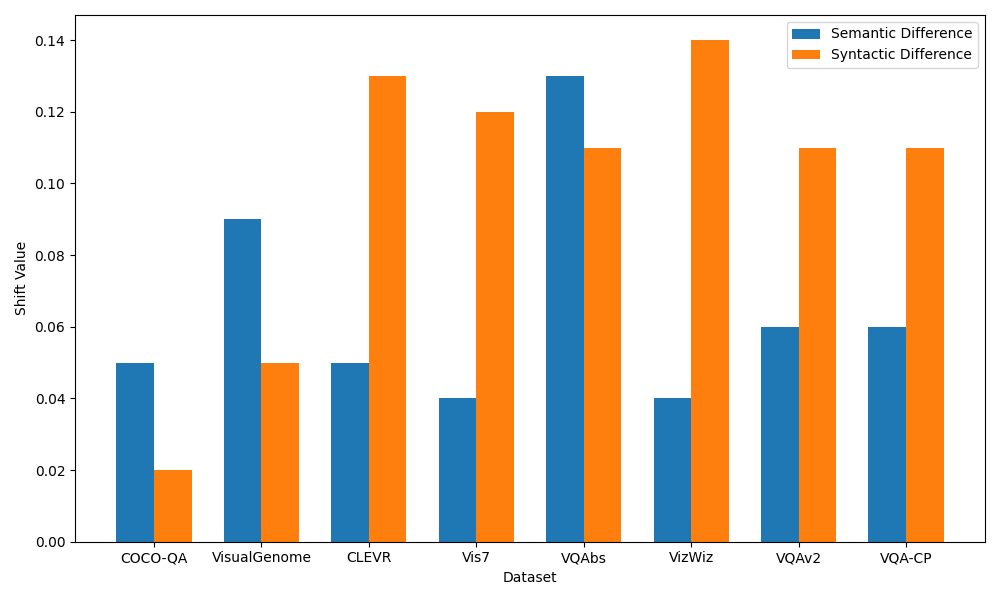}
    \caption{Syntactic and semantic shifts for question distributions. }
    \label{fig:question-shift}
  \end{subfigure}
  \caption{Domain shift analysis between VQA-GEN and other VQA datasets.}
  \label{fig:domain-shift}
\end{figure*}
\begin{figure}
\centering
\captionsetup{skip=0pt}
\includegraphics[scale=0.5]{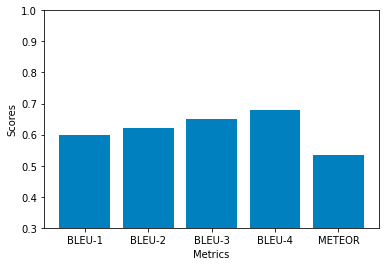}
\caption{Bar chart illustrating the evaluation of question relevance between VQA-GEN and VQA v2 using metrics BLEU-1, BLEU-2, BLEU-3, BLEU-4 and METEOR}
\end{figure}
\subsection{Quality of VQA-GEN dataset}
In this section, we analyze the quality of the VQA-GEN dataset using standard metrics and also compare the domain shift gap between other VQA datasets. Figure 5 shows how the image and question of existing VQA datasets differ from VQA-GEN dataset. We compared VQA-GEN with eight well-established benchmark datasets, namely COCO-QA \cite{ren2015exploring}, Visual Genome \cite{krishna2016visual}, CLEVR \cite{johnson2016clevr}, Visual7W \cite{zhu2016visual7w}, VQA Abs \cite{agrawal2016vqa}, VizWiz \cite{gurari2018vizwiz}, VQA v2, and VQA-CP. 
The findings demonstrate that although VQA-GEN is derived from VQA v2, it exhibits different distribution patterns in terms of both image (0.025, 0.0038) and text (0.06, 0.11) compared to VQA v2. This indicates that VQA-GEN differs from the original VQA v2 dataset. Regarding image space, VQA Abstract shows the highest distribution gap with VQA-GEN (0.05, 0.07), while VQA v2 and VQA-CP exhibit the lowest gap.\\
Similarly, in the question space, the syntactic shift is highest for VizWiz (0.14) and lowest for COCO-QA (0.02). In terms of semantic shift, VQA Abs leads with a score of 0.13, while Visual7W shows the least shift at 0.04. We observe that VQA-GEN introduces a domain gap when compared to other benchmark VQA datasets while maintaining an unbiased nature towards any specific dataset.\\
To further validate the quality of VQA-GEN, we performed a human evaluation study (details in supplementary) which showed the generated image-question pairs maintain high relevance based on human judgements.
Moving on to image and question context, Figure 6 presents a comparison of evaluations for questions' relevance using BLEU and METEOR scores. BLEU-1, BLEU-2, BLEU-3, and BLEU-4 scores range from 0.6 to 0.7, indicating substantial context similarity with the original questions. Additionally, the METEOR score (0.533900) suggests that the generated questions capture a broad spectrum of semantic variations. For image diversity, we evaluated the VQA-GEN dataset using two standard metrics - Inception Score and Fréchet Inception Distance (FID). The Inception Score is 111.49 (higher is better), indicating high image diversity. The FID score of 0.50 versus VQA v2 suggests strong contextual consistency between datasets. From the analyses, we observe that VQA-GEN introduces greater question and image diversity while maintaining relevance for VQA, addressing RQ3.
\section{Conclusion}
In this paper, we make an attempt to solve the lack of comprehensive benchmark datasets for VQA and propose VQA-GEN, the first multi-modal dataset for distribution shifts incorporating systematic and aligned shifts across visual and textual domains. The dataset was generated via an automated pipeline incorporating techniques like style transfer and backward translation to introduce visual and textual variations while preserving contextual coherence. Our dataset exposes fragility in existing VQA models reliant on unimodal distributions. Key results confirmed that textual variations alone are insufficient - truly robust VQA requires expansions in both visual and textual domains. By enabling models to learn more flexible joint representations, VQA-GEN represents a useful resource for training adaptable VQA systems that can reliably perform under diverse real-world conditions. Overall, this work took a significant step towards comprehensive domain generalization for multi-modal VQA.
\bibliographystyle{IEEEtran}
\bibliography{reference}

\title{\Large Supplementary Material \relatedversion}

\date{}

\maketitle


\fancyfoot[R]{\scriptsize{Copyright \textcopyright\ 2023 by SIAM\\
Unauthorized reproduction of this article is prohibited}}





\begin{abstract} \small\baselineskip=9pt In our paper, we propose VQA-GEN, the first ever multi-modal benchmark dataset for distribution shift generated through a shift-induced pipeline. We showed that state-of-the-art VQA models struggle to generalize when exposed to VQA-GEN dataset. In this supplementary material, we elaborate on the user study experiment to evaluate the relevance of the generated image-question pairs.
\end{abstract}

\section{Human Evaluation of Image-Question Relevance}

\subsection*{Methods}

To assess the quality of the generated image-question pairs, we conducted a survey study involving human participants to evaluate relevance. We recruited 10 students from Arizona State University to take part in an online survey consisting of 10 sets, each comprising an image, an original question, a computer-generated question, and an irrelevant question. Participants were asked to rate the relevance of each question in relation to the paired image using a 5-point Likert scale (1=Not Relevant, 5=Highly Relevant).\\
The primary objectives of this evaluation were as follows:

1) To determine whether the generated image maintains relevance with the original question. This assessment validates that image manipulation alone does not alter the contextual meaning of the question.
2) To assess whether both the generated image and question remain relevant when shifted together. This evaluation aims to confirm that concurrent adjustments to both image and question preserve relevance.
3) To ascertain whether an irrelevant question is evidently irrelevant when paired with the generated image. This test verifies that image manipulation does not introduce new irrelevant elements.

\subsection{Results}

A repeated measures ANOVA analysis revealed a significant impact of question type on relevance ratings (F(2,28)=239.42, p<.001). Further planned contrasts disclosed no noteworthy difference between the relevance ratings for the original (M=3.97, see Figure 1) and generated (M=3.82) questions (t(14)=1.59, p=.134). This result supports the notion that relevance is maintained even after joint shifting of image and question. However, it was observed that generated questions (M=3.82, see Figure 2) were consistently rated as significantly more relevant than irrelevant questions (M=1.23, see Figure 3) (t(14)=20.14, p<.001), thus confirming that image manipulation did not introduce irrelevant features.\\
Qualitative feedback provided additional insights, noting that the generated and irrelevant questions exhibited structural similarities, but the irrelevant questions contained random words that altered their intent and meaning. Overall, these findings underscore the effectiveness of our joint image-question generation approach in producing contextually relevant training data that aligns with human judgments of relevance.
\begin{figure}[htbp]
    \centering
    \includegraphics[width=0.9\linewidth]{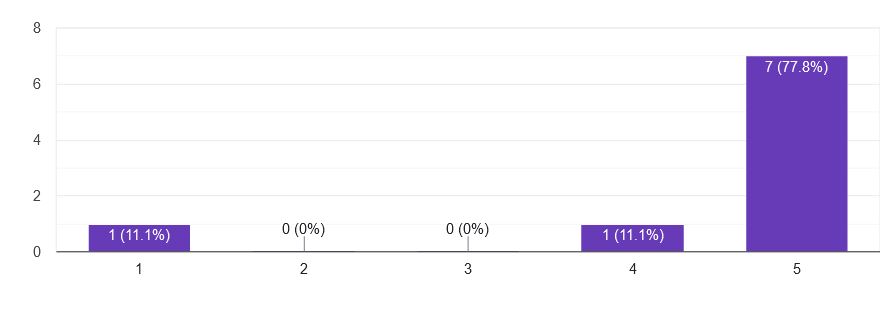}
    \caption{Relevance Ratings for Original Questions}
    \label{fig:original}
\end{figure}
\begin{figure}[htbp]
    \centering
    \includegraphics[width=0.9\linewidth]{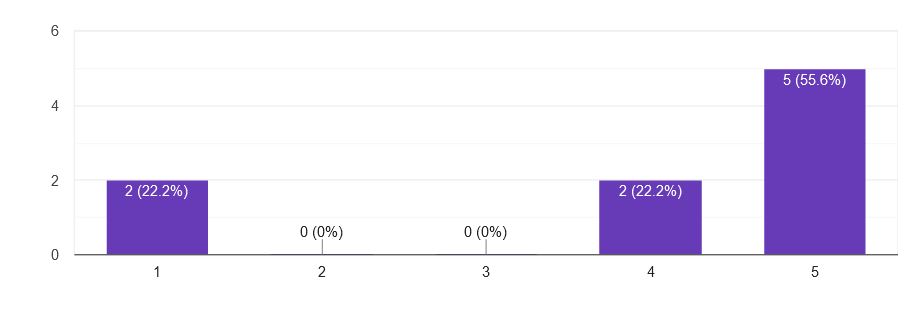}
    \caption{Relevance Ratings for Generated Questions}
    \label{fig:generated}
\end{figure}
\begin{figure}[htbp]
    \centering
    \includegraphics[width=0.9\linewidth]{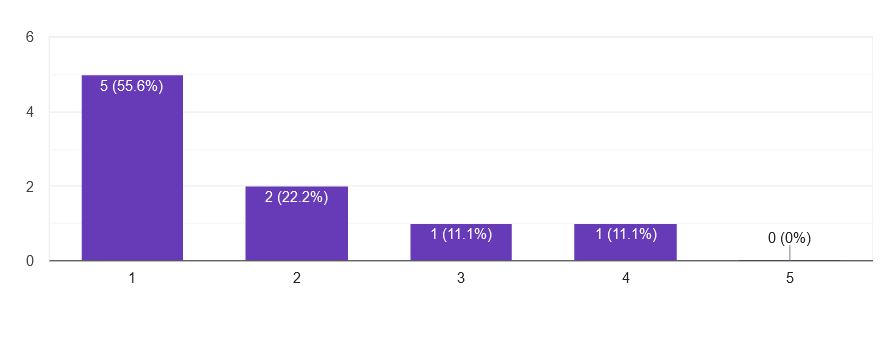}
    \caption{Relevance Ratings for Irrelevant Questions}
    \label{fig:irrelevant}
\end{figure}

\end{document}